\documentclass[sigconf, nonacm]{acmart}
\AtBeginDocument{%
  }

\copyrightyear{2026}
\acmYear{2026}
\setcopyright{cc}
\setcctype{by}
\acmConference[KDD '26]{Proceedings of the 32nd ACM SIGKDD Conference on Knowledge Discovery and Data Mining V.2}{August 09--13, 2026}{Jeju Island, Republic of Korea}
\acmBooktitle{Proceedings of the 32nd ACM SIGKDD Conference on Knowledge Discovery and Data Mining V.2 (KDD '26), August 09--13, 2026, Jeju Island, Republic of Korea}
\acmDOI{10.1145/3770855.3817490}
\acmISBN{979-8-4007-2259-2/2026/08}

\begin{document}

\title{CausalFlip: A Benchmark for LLM Causal Judgment Beyond Semantic Matching}

\author{Yuzhe Wang}
\affiliation{%
  \institution{University of Virginia}
  \city{Charlottesville}
  \state{Virginia}
  \country{USA}
}
\email{srq3da@virginia.edu}

\author{Yaochen Zhu}
\affiliation{%
  \institution{University of Virginia}
  \city{Charlottesville}
  \state{Virginia}
  \country{USA}
}
\email{uqp4qh@virginia.edu}

\author{Jundong Li}
\affiliation{%
  \institution{University of Virginia}
  \city{Charlottesville}
  \state{Virginia}
  \country{USA}
}
\email{jundong@virginia.edu}


\begin{abstract}
As large language models (LLMs) witness increasing deployment in complex, high-stakes decision-making scenarios, it becomes imperative to ground their reasoning in causality rather than spurious correlations. However, strong performance on traditional reasoning benchmarks does not guarantee true causal reasoning ability of LLMs, as high accuracy may still arise from memorizing semantic patterns instead of analyzing the underlying true causal structures. To bridge this critical gap, we propose a new causal reasoning benchmark, \textbf{CausalFlip}, designed to encourage the development of new LLM paradigms or training algorithms that ground LLM reasoning in causality rather than semantic correlation. CausalFlip consists of causal judgment questions built over event triples that could form different confounder, chain, and collider relations. Based on this, for each event triple, we construct pairs of semantically similar questions that reuse the same events but yield opposite causal answers, where models that rely heavily on semantic matching are systematically driven toward incorrect predictions. To further probe models’ reliance on semantic patterns, we introduce a noisy-prefix evaluation that prepends causally irrelevant text before intermediate causal reasoning steps without altering the underlying causal relations or the logic of the reasoning process. We evaluate LLMs under multiple training paradigms, including answer-only training, explicit Chain-of-Thought (CoT) supervision, and a proposed internalized causal reasoning approach that aims to mitigate explicit reliance on correlation in the reasoning process. Our results show that explicit CoT can still be misled by spurious semantic correlations, where internalizing reasoning steps yields substantially improved causal grounding, suggesting that it is promising to better elicit the latent causal reasoning capabilities of base LLMs. The code and benchmark is available at \url{https://github.com/Yuzhe-W/CausalFlip}.

\end{abstract}

\maketitle
\section{Introduction}
\label{sec:1}

\begin{figure}[t]
  \centering
  \includegraphics[width=\linewidth]{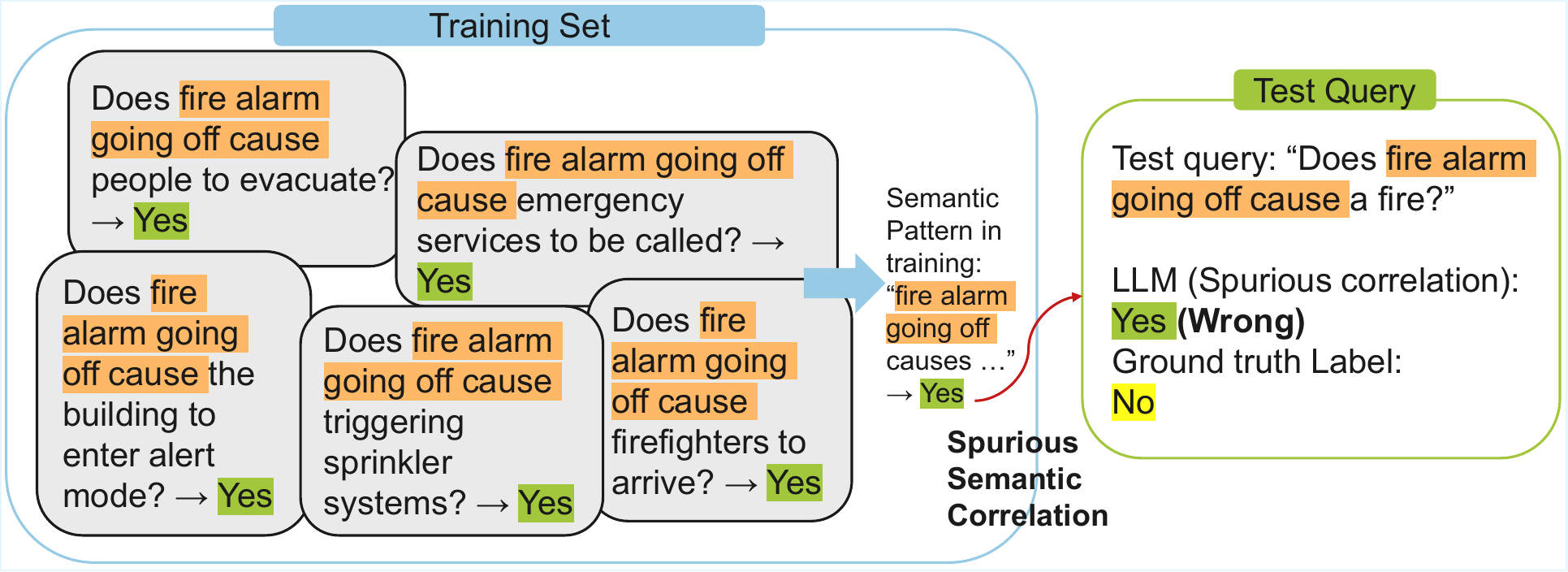}
  \caption{One representative example where training samples may create a spurious semantic correlation with a wrong answer that leads LLM to an incorrect causal judgment.}
  \label{fig:open}
  \vspace{-3mm}
\end{figure}

\noindent With the unprecedented knowledge and reasoning ability of large language models (LLM), recent years have witnessed an increased deployment in critical domains, such as medical diagnosis \cite{ApplyInMedicalUse, ApplyInMedicalUse2}, financial analysis \cite{ApplyInFinance1, ApplyInFinance2}, and legal systems \cite{ApplyInLegal}. However, these LLMs are often questioned in terms of their reliability in reasoning and decisions, where hallucinations could cause devastating outcomes \cite{hallucination1, hallucination2, hallucination3}. Traditional auto-regressive LLMs \cite{AutoLimitaions,AutoRGPT1, AutoRGPT3} optimize the model by predicting the next token \textit{conditioning on} the context, which essentially leverage semantic patterns that correlate with the training data to reason and answer for new questions \cite{Shortcuts1, Shortcuts2} (see Fig. \ref{fig:open}).
In addition, traditional reasoning benchmarks \cite{COPA,BigBench, BigBenchHard,ropesCausal} may not reliably reveal the causal reasoning limitations of LLMs, as models can often achieve high scores by exploiting semantic correlations rather than engaging in deeper causal reasoning.
Therefore, a significant gap exists between traditional benchmarks and the causal reasoning ability of LLMs, which severely hinders the development of new LLM paradigms or training algorithms that fundamentally ground their reasoning in causality.

In the traditional paradigm of auto-regressive LLMs, chain–of-\
thought (CoT) prompting may appear as the closest form of strategy to improve LLM performance on causal reasoning tasks \cite{ChainOfThoughtWei, ChainOfThoughtZeroshot}, where the later reasoning steps and the final answer "causally" depend on the previous reasoning steps. On traditional reasoning benchmarks, explicitly encouraging the LLM to generate intermediate reasoning steps often improves accuracy and makes predictions easier to inspect. However, explicit CoT leads to higher latency and token usage \cite{CoTCost, CoTCost2}. Furthermore, in essence, the CoT structures themselves can be viewed as modeling the conditional distribution of the next reasoning steps based on the context of the question and previous reasoning steps, which still rely on semantic patterns for the model to memorize and not causal. Recent work of implicit CoT proposes a different approach \cite{deng2024implicitcot}: for math tasks like multi–digit multiplication, the method gradually removes CoT steps during the model's training, nudging the model to encode the CoT in its internal weights, which allows the LLMs to solve these math tasks with high accuracy. However, it remains unclear whether such internalization better encourages the model to internalize causal reasoning model weights during the forward propagation.

\begin{figure*}[t]
  \centering
  \includegraphics[width=\textwidth]{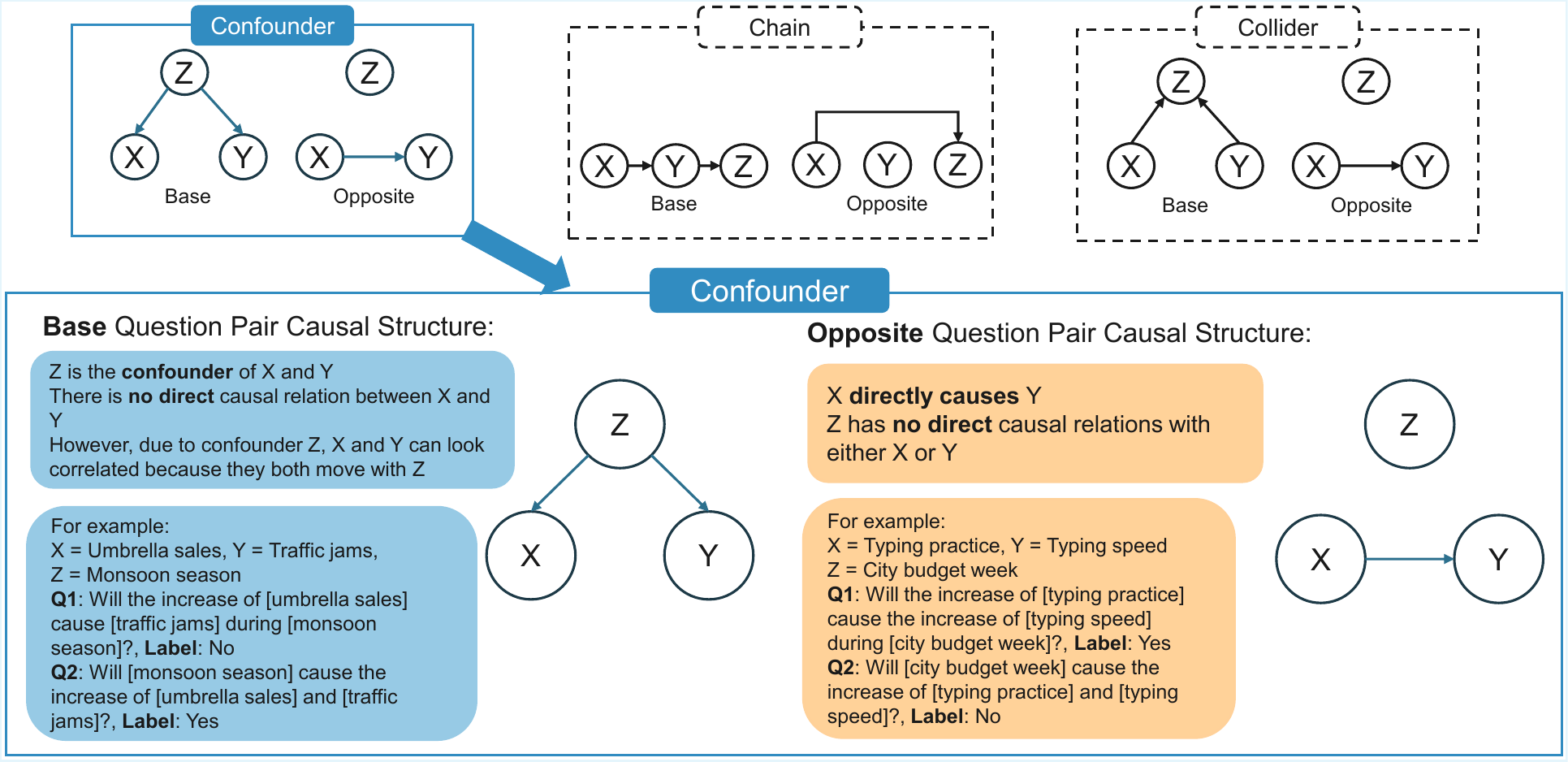}
  \caption{Overview of the causal structures used in our benchmark and an example of base, opposite pairs. The top row shows the causal structures of three sub-datasets (confounder, chain, collider); The bottom expands the confounder case and highlights a base vs. opposite question pairs: In both base and opposite question pairs, Q1 asks whether the causal relation from X to Y exists under the context of Z, with fixed template: \texttt{Will the increase of X cause Y during Z?}, and Q2 asks whether the causal relations between Z and X / Y exists, with fixed template: \texttt{Will Z cause the increase of X and Y}.}
  \label{fig:causal-pairs}
  \vspace{-4mm}
\end{figure*}

In this paper, we pioneer to bridge the gap by establishing a novel benchmark, i.e., \textbf{CausalFlip}, aiming to encourage the development of new LLM paradigms or training algorithms that fundamentally ground their reasoning in causality. The overarching principle is that LLMs that are designed or trained to generate answers based on semantic similarity (e.g., with auto-regressive next token prediction) will suffer from spurious semantic correlation, i.e., the correlation between the semantics in the question and the \textit{wrong} causal answer, whereas the models that can truly leverage causality to reason prevail. Shown in top of Figure \ref{fig:causal-pairs}, our benchmark has three sub-datasets, i.e., \textbf{confounder}, \textbf{chain}, and \textbf{collider}. Each sub-dataset includes causal questions with two causal structures: \textit{base} and \textit{opposite}. Specifically, \emph{Base} refers to the case where the event in the question contains a causal structure consistent with the sub-dataset name (e.g., in the confounder dataset, the base causal structure refers to questions where included events form the confounder structure), whereas \emph{Opposite} denotes the question with events that form a different causal structure. For each structure, we collected paired causal inquiries, where two semantically similar questions are created based on the same events with different labels (Although the event set could be different for the Base/Opposite structure). As the \emph{base} pair example shown in Figure \ref{fig:causal-pairs}'s \textbf{blue part}: Q1 and Q2 form two types of causal questions with the same event triple (Umbrella sales, Traffic jams, Monsoon season), and they have semantically similar fixed templates, specified in the caption of Figure \ref{fig:causal-pairs}. From the \emph{base} causal structure (confounder), the pairwise questions have different labels. Therefore, when we split Q1/Q2 into the training/test set, it induces spurious semantic correlation between the question semantic and the label, where LLMs that rely on semantic correlation will fail. For pairs in the \emph{opposite} structure (Figure \ref{fig:causal-pairs}'s \textbf{orange part}), Q1 and Q2 ask the same two types of causal questions over different event triples (Typing practice, Typing speed, City budget week), and use the same templates as in the \emph{base} question pair. But under the \textit{opposite} causal structure, the question that has the same template as the \emph{base} structure has the opposite answer. This prevents possible positive spurious correlations (i.e., the semantic patterns in the question that correlate with the correct label) from only including the \emph{base} structure in each sub-dataset, where Q1's labels are always ``No" and Q2's labels are always ``Yes", and the model could rely on the specific question template to predict the label. 

To further reduce possible positive spurious correlation of answer on any phrasing pattern in the inquires, two templates, \textit{Default} (default phrasing) and Alternative (an alternative phrasing grounded in the same causal structure) are provided, which leads to four categories: Base–Default (BD), Base–Alternative (BA), Opposite–Default (OD), and Opposite–Alternative (OA). Each category has the same number of causal question pairs. Finally, we use a pairwise train-test split: for each pair, one question is used for training, and its semantically similar counterpart with the opposite label is held out for evaluation, so approaches that rely heavily on semantic pattern matching are systematically penalized, and strong performance needs to be grounded in causal structure.

As an initial exploration of causality-grounded LLM, inspired by implicit CoT \cite{deng2024implicitcot}, we adopt and test \textbf{\textit{implicit causal reasoning}}, a LLM training strategy that internalizes causal reasoning steps by progressively masking the supervised intermediate causal reasoning step tokens, as a strong baseline on the \textbf{CausalFlip} dataset. Under the progressive token mask, the model is encouraged to encode the causal reasoning in its parameters rather than depending on explicit generation steps. To further investigate whether implicit causal reasoning encourages the model to focus more on causal reasoning rather than semantics, we further extend the \textbf{CausalFlip} dataset by prepending a generic, causally irrelevant prefix before the CoT as semantic noise, while keeping the CoT content and the underlying causal structure unchanged. If explicit CoT relies on semantic patterns instead of causality for reasoning, it should suffer more from the injected noise. In contrast, implicit CoT's token-removal schedule may teach the model to ``ignore'' such prefixes, where the explicit spurious correlation on generated reasoning gaps can be alleviated.

The extensive experiments conducted on the \textbf{CausalFlip} benchmark highlight two main findings. First, the limited performance of the standard LLM training strategy (no-CoT) underscores that success on CausalFlip requires grounding predictions in causal structure, rather than relying on spurious semantic correlations. In addition, the supervision of intermediate causal reasoning steps (explicit-CoT and implicit causal reasoning) improves accuracy on causal tasks compared to the pretraining strategy baseline and no-CoT fine-tuning. This suggests the intermediate causal reasoning steps as a main driver of better causal-judgment performance on CausalFlip, indicating that answering these questions needs reasoning grounded in causal structure. Second, when a fixed noisy prefix is added without changing the underlying causal structure, explicit-CoT exhibits a larger performance degradation, while implicit causal reasoning demonstrates robustness and achieves higher accuracy than explicit-CoT. This suggests that internalizing the reasoning process can reduce reliance on spurious semantic correlations on explicitly generated intermediate causal reasoning steps and better preserve decisions to causal questions. The contribution of this work can be summarized in three-fold as:
\vspace{-0.1in}
\begin{itemize}
  \item We introduce \textbf{CausalFlip}, a causality-grounding benchmark consisting of paired, semantically similar questions that involve the same events but receive opposite labels. This design aims to stimulate the development and training of LLMs whose reasoning is explicitly anchored in causal structure rather than superficial semantic cues.
  \item We propose an \textit{implicit causal reasoning} training strategy, inspired by implicit CoT \cite{deng2024implicitcot}, that progressively weakens supervision on intermediate reasoning tokens to encourage internalization, and we show it is consistently more robust than explicit-CoT under noisy-prefix distraction, indicating reduced dependence on spurious semantic correlations towards causal tasks.
  \item We introduce a novel \textit{noisy-prefix evaluation} to test whether training strategies encourage causal reasoning rather than relying on semantic pattern matching as with traditional auto-regressive language modeling.
\end{itemize}

\section{Preliminaries}
\label{sec:2}
\subsection{LLM Basics}
\label{sec:2.1}
We follow the standard language modeling formulation and view LLM as a parameterized probabilistic model over sequences of discrete tokens \cite{ProbLM,AutoRGPT1,AutoRGPT3}.
Let $\mathcal{V}$ denote the vocabulary set of the LLM and let a tokenized sequence be $\mathbf{x}=(x_1,\ldots,x_T)$ with $x_t\in\mathcal{V}$.
An autoregressive LLM with parameters $\theta$ defines a joint distribution over $\mathbf{x}$ via the chain-rule factorization as follows:
\begin{equation}
p_\theta(\mathbf{x}) \;=\; \prod_{t=1}^{T} p_\theta(x_t \mid x_{<t}),
\label{eq:ar-factorization}
\end{equation}
where $x_{<t} = (x_1, \ldots, x_{t-1})$ is the prefix context.
Autoregressive generation produces tokens sequentially from left to right by decoding $x_t \sim p_\theta(\cdot \mid x_{<t})$.
In practice, $p_\theta(x_t \mid x_{<t})$ is commonly instantiated by a Transformer architecture based on self-attention \cite{Attention}.
Concretely, let $h_t$ denote the final-layer hidden representation at the last position of the prefix $x_{<t}$.
The next-token distribution is then obtained by applying a softmax over vocabulary logits,
\begin{equation}
p_\theta(x_t=v \mid x_{<t}) \;=\; \frac{\exp(s_\theta(v; h_t))}{\sum_{v'\in\mathcal{V}} \exp(s_\theta(v'; h_t))},
\label{eq:softmax}
\end{equation}
where $s_\theta(\cdot;h_t)$ denotes the score (logit) assigned to each vocabulary token. In generation, the model conditions on a given context $\mathbf{c}=(c_1,\ldots,c_K)$ and generates a continuation $\mathbf{y}=(y_1,\ldots,y_M)$.
The induced conditional distribution follows the same autoregression:
\begin{equation}
p_\theta(\mathbf{y}\mid \mathbf{c})
\;=\;
\prod_{m=1}^{M} p_\theta(y_m \mid \mathbf{c}, y_{<m}),
\label{eq:conditional-generation}
\end{equation}
where $y_{<m}=(y_1,\ldots,y_{m-1})$.
The standard training objective of LLMs is the maximum likelihood estimation over a corpus $\mathcal{D}$ (equivalently, minimizing token-level cross-entropy):
\begin{equation}
\max_{\theta}\; \sum_{\mathbf{x}\in \mathcal{D}} \sum_{t=1}^{T} \log p_\theta(x_t \mid x_{<t}).
\label{eq:mle}
\end{equation}
This next-token prediction objective rewards any statistical regularities in the context that improve predictive likelihood, which can include spurious semantic correlations with downstream labels \cite{AutoRGPT3}.

At inference time, a completion is generated by decoding from Eq.~\eqref{eq:conditional-generation}, for example via greedy decoding, or stochastic sampling \cite{sqtosq,topP}.
In particular, sampling-based decoding, e.g., top-$k$, is commonly used to trade off diversity and coherence in the generated text \cite{topK}. From Eq. \eqref{eq:conditional-generation} we can find that, the training of LLMs with auto-regressive language modeling relies on conditional distribution instead of causal analysis on the questions, which is susceptible to provide the wrong answers based on spurious semantic correlations.

\subsection{Chain-of-Thought Reasoning}
\label{sec:2.2}

In this part, we formalize explicit Chain-of-Thought (CoT), which provides the foundation for the proposed implicit causal reasoning and the robustness evaluation. CoT reasoning refers to generating intermediate natural-language reasoning steps before producing a final answer, which has been shown to improve performance on multi-step reasoning tasks \cite{ChainOfThoughtWei,ChainOfThoughtZeroshot}.
Formally, let the input/prompt be $\mathbf{x}$. Under CoT, the model's output is a single sequence $\mathbf{z}$, which we split into \textit{\textbf{(i)}} intermediate reasoning steps (CoT) tokens $\mathbf{s}$ and \textit{\textbf{(ii)}} final answer tokens $\mathbf{y}$ as $\mathbf{z} = [\mathbf{s}; \mathbf{y}]$.
An autoregressive language model defines the conditional likelihood of the full output as
\begin{equation}
p_\theta(\mathbf{z}\mid \mathbf{x})
\;=\;
\prod_{t=1}^{|\mathbf{z}|} p_\theta(z_t \mid \mathbf{x}, \mathbf{z}_{<t}),
\label{eq:cot-ar-z}
\end{equation}
where $\mathbf{z}_{<t}$ is the prefix of previously generated output tokens.
By viewing $\mathbf{z}$ as $\mathbf{z} = [\mathbf{s}; \mathbf{y}]$, it is equivalent to:
\begin{equation}
p_\theta(\mathbf{s}, \mathbf{y}\mid \mathbf{x})
\;=\;
p_\theta(\mathbf{s}\mid \mathbf{x})\; p_\theta(\mathbf{y}\mid \mathbf{x}, \mathbf{s}).
\label{eq:cot-factorize}
\end{equation}
From this perspective, $\mathbf{s}$ serves as an intermediate reasoning steps which the final decision $\mathbf{y}$ is conditioned \cite{ChainOfThoughtWei}. At inference time, CoT typically proceeds by generating a \emph{single} reasoning step's rollout $\hat{\mathbf{s}}$ via a decoding procedure $\mathrm{Dec}(.)$ (e.g., greedy decoding or sampling), then predicting the answer $\hat{\mathbf{y}}$ conditioned on it:
\begin{equation}
\hat{\mathbf{s}} = \mathrm{Dec}_\theta(\mathbf{x}),
\ 
\hat{\mathbf{y}} = \mathrm{Dec}_\theta([\mathbf{x}; \hat{\mathbf{s}}]).
\label{eq:cot-rollout}
\end{equation}
 However, CoT is essentially correlational instead of causal, as each intermediate step is still based on the conditional distribution given the question and previously generated reasoning steps, which therefore relies on spurious semantic correlation instead of grounding the reasoning in the true underlying causal structure. 
\subsection{Problem Formulation}
\label{sec:2.3}
To address the limitation of traditional auto-regressive LLM and explicit CoT, we study causal-judgment question answering with LLMs.
Let $\mathcal{Q}$ denote the space of natural-language causal questions, and let $\mathcal{A}$ denote the answer space, where $\mathcal{A}=\{\texttt{Yes},\texttt{No}\}$ denotes the binary causal answers.
We assume each question $q\in\mathcal{Q}$ is associated with an underlying causal graph $G$ and a target causal label $a\in\mathcal{A}$ that reflects the correct causal judgment under $G$. We represent the benchmark as a dataset $\mathcal{D}=\{(q_i, G_i, a_i)\}_{i=1}^{N}$. Here, we assume a general formulation of LLM with parameters $\theta$, which defines the probability of the answer (in textual format) given the question $
p_\theta(a \mid q), a\in\mathcal{A}$, 
and predicts an answer by $\hat{a}(q) = \arg\max_{a\in\mathcal{A}} p_\theta(a \mid q)$.

\section{Benchmark Design}
\label{sec:3}
The aim of this benchmark is to address the limitations of traditional benchmarks in comprehensively evaluating LLMs’ ability to make reliable causal judgments. CausalFlip is built from event triples $(X,Y,Z)$ and formulates binary causal judgment questions across three sub-datasets: \emph{confounder}, \emph{chain}, and \emph{collider}. For each sub-dataset, we define \emph{Base} and \emph{Opposite} causal structures and construct semantically similar question pairs whose correct labels flip according to the underlying causal structure. Combined with a pairwise train-test split, this design penalizes models from relying on spurious semantic correlations for predictions. Furthermore, we control question templates through Default and Alternative templates to reduce template-driven positive spurious correlation.

\subsection{Semantically Similar, Label-Flipped Pairs with Pairwise Train–Test Split}
\label{sec:3.1}
The overarching principle for developing the \textbf{CausalFlip} dataset is to introduce \textit{paired questions with spurious correlations in semantics, i.e., questions with similar semantic meanings but lead to different causal answers}. The two questions differ primarily in the causal relation they query, and therefore have opposite labels, corresponding to the two question types \textbf{\textit{(i)}}, \textbf{\textit{(ii)}} defined in subsection~\ref{sec:3.2}. For example, in the confounder dataset, a pair contrasts a direct-effect question, \texttt{Will the increase of $X$ cause $Y$ during $Z$?}, with a semantically similar confounder question, \texttt{Will $Z$ cause the increase of $X$ and $Y$?}. For evaluation, we adopt a \emph{pairwise} train--test split. In each dataset, for each pair, we assign one question to the training set and the other to the test set while keeping the number of Q1/Q2 balanced across the training and test splits. This ensures that every test question has a counterpart in training that shares the same event triples and similar semantics but carries the opposite label. Consequently, if the model relies heavily on semantic patterns, it will predict the same label as the corresponding training instance, thereby failing on the paired testing question where the correct label is flipped. Thus, a high performance on this benchmark requires the understanding of causal structures, rather than relying on semantic matching.

\subsection{Causal Structures and Induced Questions}
\label{sec:3.2}
The established \textbf{CausalFlip} benchmark is composed of three sub-datasets, i.e., \emph{confounder}, \emph{chain}, and \emph{collider}, where each dataset type studies the canonical causal structure among $(X, Y, Z)$. Specifically, within each dataset type, we define two causal structures: \emph{Base (B)} and \emph{Opposite (O)}. \emph{Base} follows the canonical causal graph of the dataset type, while \emph{Opposite} uses an alternative causal graph that is constructed to yield the opposite labels to the same paired question types compared to the \emph{Base}. With this design, the model cannot rely on the positive spurious correlation between the question template and the label (e.g., with only Base pairs, question 1's label is always ``No" and question 2's is always Yes). This forces correct predictions to depend on the underlying causal structures. To keep the benchmark balanced, we also construct the same number of pairs under the \emph{Base} and \emph{Opposite} causal structures in each dataset.

\vspace{2mm}
\subsubsection{\textbf{Confounder Dataset.}}
\textbf{Base (B)} represents the confounder \textit{base} structure, where $Z \rightarrow X$ and $Z \rightarrow Y$, with no directed causal path from $X$ to $Y$. Each question pair in this dataset includes two types of questions: \textbf{\textit{(i)}} is a direct-effect question asking whether increasing $X$ causes $Y$ \emph{in the context of} $Z$ (we include $Z$ as context to keep this question semantically close to question~\textbf{\textit{(ii)}}), and \textbf{\textit{(ii)}} is a confounder question asking whether $Z$ causes both $X$ and $Y$. Under the \emph{Base} structure, Question~\textbf{\textit{(i)}} is labeled ``\texttt{No}" and Question~\textbf{\textit{(ii)}} is labeled ``\texttt{Yes}".
\textbf{Opposite (O)} represents the \textit{opposite} structure under confounder dataset, that follows $X \rightarrow Y$, and there is no causal path from $Z$ to both $X$ and $Y$, which means $Z$ is non-causal w.r.t. both $X$ and $Y$. The question types \textit{i}, \textit{ii} remain the same, but due to the opposite causal structure, the labels flip: Question~\textbf{\textit{(i)}} is assigned \texttt{Yes} and Question~\textbf{\textit{(ii)}} is assigned \texttt{No}.

\subsubsection{\textbf{Chain Dataset.}}
\textbf{Base (B)} represents the chain \textit{base} structure follows $X \rightarrow Y \rightarrow Z$, with no direct causal path $X \rightarrow Z$, which means any effect of $X$ on $Z$ is mediated through $Y$. Each question pair in this dataset includes two question types: \textbf{\textit{(i)}} is a direct-effect question asking whether increasing $X$ directly causes $Z$ \emph{in the context of} $Y$ (we include $Y$ as context to keep this question semantically close to question~\textbf{\textit{(ii)}})), and \textbf{\textit{(ii)}} is a mediated path question asking whether increasing $X$ increases $Y$, which in turn increases $Z$. Under Base structure, Question~\textbf{\textit{(i)}} is labeled \texttt{No} and Question~\textbf{\textit{(ii)}} is labeled \texttt{Yes}.
\textbf{Opposite (O)} represents \textit{opposite} structure under chain dataset, follows $X \rightarrow Z$, and there is no causal path from $X$ to $Y$ and $Y$ to $Z$. The question types remain the same, but due to the opposite causal structure, labels flip: Question~\textbf{\textit{(i)}} is assigned \texttt{Yes} and Question~\textbf{\textit{(ii)}} is assigned \texttt{No}.

\subsubsection{\textbf{Collider Dataset.}}
\textbf{Base (B)} represents the collider \textit{base} structure follows $X \rightarrow Z$ and $Y \rightarrow Z$, with no directed path between $X$ and $Y$; $Z$ is a collider with two independent parents. Each instance includes two question types: \textbf{\textit{(i)}} is a direct-effect question asking whether increasing $X$ causes $Y$ \emph{in the context of} $Z$ (we include $Z$ as context to keep this question semantically close to Question~\textbf{\textit{(ii)}}), and \textbf{\textit{(ii)}} is a collider question asking whether increasing $X$ causes $Z$ and increasing $Y$ causes $Z$. Under Base structure, Question~\textbf{\textit{(i)}} is labeled \texttt{No} and Question~\textbf{\textit{(ii)}} is labeled \texttt{Yes}.
\textbf{Opposite (O)} represents the \textit{opposite} structure under collider dataset, follows $X \rightarrow Y$, and there is no causal path from $X$ to $Z$ and no causal path from $Y$ to $Z$. The question types remain the same, but labels flip: Question~\textbf{\textit{(i)}} is assigned \texttt{Yes} and Question~\textbf{\textit{(ii)}} is assigned \texttt{No}.

\subsection{Question Templates}
\label{sec:3.3}

The design of pairwise split introduces a strong spurious (negative) correlation between question semantics and causal answers, which ideally should lead to the failure of LLM training strategies that purely rely on semantic correlation to generate answers. However, we find that the auto-regressive LLMs may still exploit shortcuts tied to phrase patterns in the question, i.e., certain remaining \textit{positive} semantics correlation with the answer. To further reduce such shortcuts on wording style, we construct two question phrasing templates as randomization to reduce the positive spurious correlation: \textbf{Default (D)} and \textbf{Alternative (A)}.
\begin{itemize}
\item \textbf{Default (D)} is default question phrasing in \textbf{CausalFlip}. It expresses each dataset's two question types as direct binary questions (e.g., \texttt{Will the increase in $X$ cause $Y$ during $Z$\ldots ?}). For each dataset type, confounder, chain, and collider, we generate a set of paired questions using the Default templates.
\item \textbf{Alternative (A)} is a different phrasing to express the same types of causal questions. It is a declarative form (e.g., \texttt{An increase in $X$ leads to an increase in $Y$ \ldots}), and the model judges whether the statement is correct with a binary \texttt{\{Yes, No\}} answer. For each dataset type, we generate a separate set of paired questions using the Alternative templates. Within each dataset, we balance the samples across template types by ensuring the same number of Default pairs and Alternative pairs. Combining template family (Default vs Alternative) with causal structure (Base vs Opposite) yields four categories: \textbf{BD}, \textbf{BA}, \textbf{OD}, and \textbf{OA}. 
\end{itemize}
Since each dataset contains the same number of Base and Opposite pairs and the same number of Default and Alternative pairs, the number of samples in BD, BA, OD, and OA is matched. This balance prevents any structure-template category (BD/BA/OD/OA) from being over-represented, and helps ensure that predictions are grounded in causal structure rather than template shortcuts.

\section{Implicit Causal Reasoning}
\label{sec:4}
In the previous section, we have proposed  \textbf{CausalFlip} benchmark that introduces spurious semantic correlation between \emph{causal question semantics} and \emph{labels} such that a model trained based on \emph{semantic pattern matching} fails, encouraging \emph{training strategies} to reason on \emph{causal structures}. In this section, we apply the strategy, implicit causal reasoning, to reduce the explicit-CoT's reliance on spurious, explicit semantic correlations in causal questions.

\subsection{Motivation and Overview}
\label{sec:4.1}
CausalFlip is designed to penalize the performance of models that rely on semantics for prediction. This motivates training strategies that reduce a model's tendency to rely on semantic correlation when making causal judgments. As a preliminary exploration in fundamentally grounding the LLM reasoning in causality, we adopt a \emph{progressive causal reasoning steps mask} strategy inspired by implicit CoT \cite{deng2024implicitcot}, where we progressively mask the causal reasoning steps, removing them from the training loss to specifically mitigate its reliance on spurious semantic correlations in explicitly generated causal reasoning steps and encourage generations grounded in causal structure.

\subsection{Progressive Causal Reasoning Steps Mask}
\label{sec:4.2}
In this part, we consider training samples consisting of an input question $\mathbf{x}$, a binary answer $y \in \{\texttt{Yes}, \texttt{No}\}$, and explicit causal reasoning steps $s=(s_1,\dots,s_L)$. In our setting, $s$ is the intermediate causal reasoning steps derived from the corresponding causal graph, which states the key structural conditions and leads to the answer.

\subsubsection{\textbf{Explicit Causal Reasoning Steps Supervision.}}
We start from explicit CoT, where all causal reasoning steps' tokens are included in the training loss. The standard autoregressive next-token objective is applied on all intermediate causal reasoning steps' tokens $s=(s_1,\dots,s_L)$ as well as the final answer $y$ as:
\begin{equation}
\label{eq:explicit}
\mathcal{L}_{\text{explicit}}(\theta)
=
-\sum_{(\mathbf{x},\mathbf{s},y)\in\mathcal{D}} \log p_\theta(\mathbf{s},y\mid \mathbf{x}),
\end{equation}
Equivalently, the token-wise likelihood is given by Eq. \eqref{eq:explicitToken}. Here $L$ denotes the number of tokens in the causal reasoning steps, and $k\in\{1,\dots,L\}$ indexes a token position within $s$:
\begin{equation}
\label{eq:explicitToken}
\begin{aligned}
\mathcal{L}_{\text{explicit}}(\theta)
&=
-\sum_{(\mathbf{x},\mathbf{s},y)\in\mathcal{D}}
\Biggl(
\sum_{k=1}^{L}\log p_\theta(s_k\mid \mathbf{x},\mathbf{s}_{<k})
+
\log p_\theta(y\mid \mathbf{x},\mathbf{s})
\Biggr).
\end{aligned}
\end{equation}
\subsubsection{\textbf{Progressive Causal Reasoning Steps Removal.}}
For implicit causal reasoning, we progressively mask the earliest causal reasoning steps' tokens from supervision. At training step $t$, we remove the first $r(t)$ tokens from supervision and only compute the loss on the remaining part. Formally, we apply a step-based mask function $m_k(t)\in\{0,1\}$, over causal reasoning steps token positions $k$ as follows:
\begin{equation}
\label{eq:implicit}
\begin{aligned}
\mathcal{L}_{\text{mask}}(\theta; t)
&=
-\sum_{(\mathbf{x},\mathbf{s},y)\in\mathcal{D}}
\Biggl(
\sum_{k=1}^{L} m_k(t)\,\log p_\theta(s_k\mid \mathbf{x},\mathbf{s}_{<k})
\\
&
+
\log p_\theta(y\mid \mathbf{x},\mathbf{s})
\Biggr),
\
m_k(t)=\mathbf{1}[k>r(t)].
\end{aligned}
\end{equation}
where $m_k(t)=0$ indicates that the loss on $s_k$ is masked out, and $m_k(t)=1$ indicates standard supervision (Eq. \eqref{eq:implicit}). The number of masked reasoning token $r(t)$ follows a stepwise schedule inspired by implicit CoT \cite{deng2024implicitcot}:

\begin{equation}
\label{eq:mask_schedule}
r(t)=\min\left(L,\left\lfloor \delta \cdot \frac{t}{T_{\mathrm{epoch}}}\right\rfloor+\epsilon\right),
\end{equation}
where $L$ is the number of reasoning tokens, $T_{\mathrm{epoch}}$ is the number of training steps in one epoch, $\delta$ is a hyperparameter controlling how many reasoning tokens are masked per epoch, and $\epsilon$ is a sampled offset used to smooth transitions between adjacent removal levels.

\section{Empirical Study}
In this section, we evaluate four strategies (i.e., naive pretraining strategy, no-CoT, explicit-CoT, and implicit causal reasoning fine-tuning) on our \textbf{CausalFlip} benchmark. We first find that the naive pretraining and no-CoT strategies perform poorly, confirming that LLMs need to ground their predictions in causal structure instead of semantic patterns to succeed in the CausalFlip benchmark. In addition, we find that training strategies with intermediate causal reasoning steps achieve much stronger causal-judgment performance. Finally, we further demonstrate that when injecting a noisy semantic prefix, implicit causal reasoning maintains higher robustness than explicit-CoT, indicating its less reliance on spurious semantic correlations and a reliable grounding in causal structure.

\label{sec:5}
\subsection{Research Questions}
\label{sec:5.1}
\noindent\textbf{RQ1: How do no-CoT, explicit-CoT, and implicit causal reasoning training strategies perform on our causal benchmark compared to the naive pretraining strategy?}

\noindent This question compares strategies in terms of their ability to make \emph{causal} judgments when spurious semantic correlations are not a reliable shortcut, due to the pair-wise split design in \textbf{CausalFlip}. All strategies use the same train/test split and the same evaluation protocol, and we report overall accuracy for each model.

\vspace{2mm}

\noindent\textbf{RQ2: Could the implicit causal reasoning model perform more robustly in causal reasoning tasks than explicit-CoT by reducing its reliance on semantic patterns?}

\noindent As we analyze in subsection \ref{sec:2.1}, autoregressive \emph{LLMs} are optimized to fit conditional distributions, and thus can exploit positive spurious semantic correlations to make predictions, instead of grounding in the underlying causal structure \cite{Shortcuts1, Shortcuts2}. Motivated by this, we further test whether implicit causal reasoning reduces reliance on semantic patterns by comparing it with explicit-CoT under our noisy-prefix evaluation. Specifically, we inject the same causally irrelevant prefix before intermediate causal reasoning steps and measure how each model’s accuracy changes, using performance degradation as an indicator of reliance on semantic patterns versus causal structure.

\subsection{Baseline and Training Strategies}
\label{sec:5.2}
We compare four strategies to answer the two research questions raised in this paper. Across all experiments, we use \textbf{Llama-3.2-3B-Instruct} as the base model.

\begin{itemize}
\item{\textbf{Naive pretraining strategy (No Fine-tuning)}: This is the base model without any training on our dataset \textbf{CausalFlip}. It measures the baseline pretraining's causal judgment performance under pretraining alone and serves as a reference for evaluating fine-tuning strategies (no-CoT, explicit-CoT, and implicit causal reasoning).

\item{\textbf{No-CoT fine-tuning}}: This is the standard answer-only supervised fine-tuning baseline, where the model is trained to directly predict the final answer given a causal question from \textbf{CausalFlip}.
In this setting, no intermediate causal reasoning steps are provided during training, so the supervision is applied only to the final \texttt{Yes/No} decision.

\item {\textbf{Explicit-CoT fine-tuning}}:  Different from no-CoT model, this model is trained with the full intermediate causal reasoning steps followed by the final answer, and supervision is applied to both the intermediate causal reasoning tokens and the final answer.

\item{\textbf{Implicit causal reasoning fine-tuning}}: We progressively mask an increasing prefix of intermediate reasoning tokens from the training loss while always supervising the final decision. 
This shifts prediction's reliance from explicit semantics to the underlying causal structure.}
\end{itemize}

These four training strategies enable the evaluation of how different supervision strategies influence model performance on our proposed causal question dataset.

\subsection{Training and Evaluation Setups}
\label{sec: 5.3}
During training and evaluation, all strategies use the same fixed train/test split set with the same causal questions defined in Section \ref{sec:3} to ensure fair comparisons. In the following part, we introduce the training protocol and evaluation protocol in more detail.

\subsubsection{\textbf{Training}}
Each fine-tuning instance is composed of \textit{\textbf{(i)}} the causal question from \textbf{CausalFlip} as defined earlier in Section \ref{sec:3}, and \textit{\textbf{(ii)}} the answer of the question, which is the binary answer in \texttt{\{Yes,No\}}. For \emph{no-CoT}, the instance is the same as we stated before, that is, the question followed by the answer. For \emph{explicit-CoT} and \emph{implicit causal reasoning}, we include the intermediate causal reasoning steps placed between the causal question and the answer (see subsection \ref{sec:4.2}).
For example, for \textit{base} causal structure pairs in the \textit{confounder} dataset, we employ the following causal reasoning step: 

\begin{center}
\fbox{
\begin{minipage}{0.95\linewidth}
\texttt{No directed causal path from {X} to {Y} AND adjusting for {Z} closes the backdoor between {X} and {Y}, therefore}
\end{minipage}
}
\end{center}
\noindent The final answer follows immediately.

\subsubsection{\textbf{Evaluation}}
At test time, models are presented with a question-only prompt and evaluated based on their generated responses.
Each prompt contains two parts: \textit{\textbf{(i)}} \emph{a format instruction} that constrains the final answer within the model's output as a binary decision \{\texttt{Yes}, \texttt{No}\}, which makes it amenable to parse and evaluate, even in the settings of explicit-CoT and implicit causal reasoning, where intermediate causal reasoning steps are generated before the final answer; and \textit{\textbf{(ii)}} a causal question from \textbf{CausalFlip}, defined in section ~\ref{sec:3}.
For each prompt, the model generates a single response; we extract the final answer and use it to compute task accuracy.

\subsubsection{\textbf{Metrics.}}
We score the generation based only on the final \textbf{Yes/No} answer without judging any intermediate causal reasoning steps.
Specifically, for each test example, we take the model's final answer, which is the predicted label $\hat{y}_i$, and compare it with the ground-truth label $y_i$.
Accuracy is computed as:
\begin{equation}
\mathrm{Accuracy}=
\frac{1}{N}
\sum_{i=1}^{N}
\mathbf{1}\left(\hat{y}_i = y_i\right).
\end{equation}
where $N$ is the number of test examples, $y_i \in \{\texttt{Yes},\texttt{No}\}$ is the ground-truth label, $\hat{y}_i \in \{\texttt{Yes},\texttt{No}\}$ is the model's predicted label
(final answer), and $\mathbf{1}[\cdot]$ is the indicator function.
To measure the performance across the models, we report accuracy on clean inputs and under the noisy-prefix setting described below.

\vspace{2mm}
\subsubsection{\textbf{Noisy-Prefix Evaluation.}}
\label{sec:5.3-noisy}
To test whether the model focuses on causal reasoning or semantics, we inject the noisy prefix before the intermediate causal reasoning steps in the training samples. The prefix is a fixed chunk of natural-language sentences designed to serve as a semantic distraction while remaining causally irrelevant, which does not violate the intermediate reasoning process, nor change the underlying causal structure. 
We apply the same prefix to all questions to ensure consistency. This avoids spurious correlations between different noisy prefixes' semantics and labels.

We compare explicit-CoT and implicit causal reasoning because they represent two training strategies for leveraging intermediate causal reasoning. Explicit-CoT exposes the model to full intermediate causal reasoning steps during training, which may increase reliance on spurious semantic correlations and make its final decision more sensitive to the noisy prefix. In contrast, implicit causal reasoning progressively masks causal reasoning steps' tokens over training. We evaluate both strategies under the noisy-prefix.

\subsection{Performance on the CausalFlip (RQ1)}
\label{sec:5.4}

We first answer \textbf{RQ1} by comparing four strategies' performances on CausalFlip: the naive pretraining strategy, standard answer-only fine-tuning (\textbf{no-CoT}), \textbf{explicit-CoT} fine-tuning, and \textbf{implicit causal reasoning} fine-tuning. We report accuracy on clean (without noisy prefix) inputs for \textbf{Confounder}, \textbf{Chain}, and \textbf{Collider} datasets (Table~\ref{tab:confounder}, \ref{tab:chain}, \ref{tab:collider})

\begin{table}[t]
\centering
\caption{Performance comparisons on Confounder Dataset}
\label{tab:confounder}
\resizebox{\columnwidth}{!}{%
\begin{tabular}{lccc}
\toprule
\textbf{Model} & \textbf{Accuracy} & \textbf{Correct / Total} & \textbf{Valid Yes/No} \\
\midrule
Naive pretraining strategy & 0.529 & 529 / 1000 & 1000 / 1000 \\
No CoT & 0.524 & 524 / 1000 & 1000 / 1000 \\
Explicit CoT & 0.892 & 892 / 1000 & 987  / 1000 \\
Implicit Causal Reasoning & 0.900 & 900 / 1000 & 997 / 1000 \\
\bottomrule
\end{tabular}
}
\vspace{-6pt}
\end{table}

\begin{table}[t]
\centering
\caption{Performance comparisons on Chain Dataset}
\label{tab:chain}
\resizebox{\columnwidth}{!}{%
\begin{tabular}{lccc}
\toprule
\textbf{Model} & \textbf{Accuracy} & \textbf{Correct / Total} & \textbf{Valid Yes/No} \\
\midrule
Naive pretraining strategy & 0.612 & 612 / 1000 & 1000 / 1000 \\
No CoT & 0.639 & 639 / 1000 & 1000 / 1000 \\
Explicit CoT & 0.690 & 690 / 1000 & 989 / 1000 \\
Implicit Causal Reasoning & 0.757 & 757 / 1000 & 998 / 1000 \\
\bottomrule
\end{tabular}
}
\vspace{-6pt}
\end{table}

\begin{table}[t]
\centering
\caption{Performance comparisons on Collider Dataset}
\label{tab:collider}
\resizebox{\columnwidth}{!}{%
\begin{tabular}{lccc}
\toprule
\textbf{Model} & \textbf{Accuracy} & \textbf{Correct / Total} & \textbf{Valid Yes/No} \\
\midrule
Naive pretraining strategy & 0.629 & 629 / 1000 & 1000 / 1000 \\
No CoT & 0.655 & 655 / 1000 & 1000 / 1000 \\
Explicit CoT & 0.856 & 856 / 1000 & 1000 / 1000 \\
Implicit Causal Reasoning & 0.849 & 849 / 1000 & 999 / 1000 \\
\bottomrule
\end{tabular}
}
\vspace{-6pt}
\end{table}
On all three datasets, we observe low performance of the naive pretraining and no-CoT fine-tuning strategies, as these strategies fundamentally rely on semantics correlations rather than grounding in causal structure to train the LLMs. On Confounder, naive/no-CoT stay near chance (0.529/0.524) while explicit-CoT and implicit reach 0.892/0.900; on Chain, no-CoT (0.639) provides only a small gain over naive (0.612), compared to explicit-CoT/implicit (0.690/0.757); and on Collider, no-CoT is only modestly above naive (0.655 vs. 0.629) while explicit-CoT/implicit achieve 0.856/0.849. These results show that the intermediate causal reasoning steps are the main driver of improved performance on causal judgment questions. Meanwhile, the performance of \textbf{implicit causal reasoning} remains competitive and \textbf{surpasses} explicit-CoT on the confounder and chain dataset. While explicit-CoT and implicit causal reasoning occasionally produce invalid final-answer formats, our manual inspection shows all these invalid answers fail in matching any recognized formats of Yes or No (e.g., Yes., no!), or other identifiable decision keywords. In addition, the limited performance of no-CoT fine-tuning suggests that achieving success on CausalFlip requires a training strategy that effectively grounds predictions in causal structure rather than exploiting spurious semantic correlations.

\label{sec:5.5}
\begin{figure*}[t]
  \vspace{-10pt}
  \centering
  \includegraphics[width=1\textwidth]{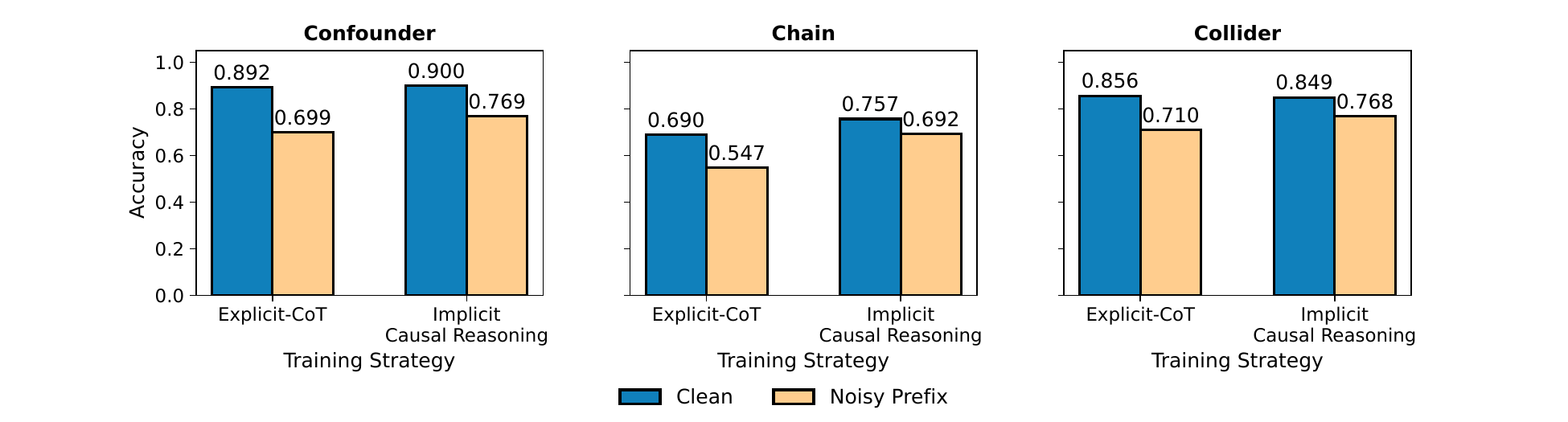}
  \vspace{-20pt}
  \caption{Accuracy of explicit-CoT versus implicit causal reasoning on CausalFlip across the three sub-datasets under clean inputs and noisy-prefix. Implicit causal reasoning consistently degrades less and performs better than explicit-CoT after the injection of noisy prefix, indicating its reduced reliance on spurious semantic correlations.}
  \label{fig:causal-noisy}
\vspace{-10pt}
\end{figure*}
\subsection{Performance under Noisy Prefix (RQ2)}
We next answer \textbf{RQ2} by evaluating explicit-CoT and implicit causal reasoning's performances on CausalFlip under the \textbf{noisy-prefix} injection (defined in section \ref{sec:5.3-noisy}). This prefix is designed to be causally irrelevant, which does not violate existing causal structure. Since the underlying causal structure is unchanged, the performance degradation indicates that the model’s prediction is sensitive to semantics, rather than being grounded in the causal structure. 

As is shown in Figure \ref{fig:causal-noisy}, across all three datasets, adding the noisy prefix consistently causes a larger accuracy drop for explicit-CoT than for implicit causal reasoning, and the average drop is about $0.161$ for explicit-CoT versus $0.092$ for implicit causal reasoning. Moreover, under the noisy-prefix setting, implicit causal reasoning achieves higher accuracy than explicit-CoT on confounder, chain, and collider ($0.769 > 0.699$, $0.692 > 0.547$, $0.768 > 0.710$). Overall, under the noisy semantic prefix setting, the larger degradation of explicit-CoT suggests that it still predicts answers by relying heavily on semantics rather than the underlying causal structure. In contrast, implicit causal reasoning maintains higher accuracy under the same distraction, indicating stronger robustness in causality-related questions and its reduced reliance on semantics.

\vspace{-5mm}
\section{Related Work}
\label{sec:6}

\noindent\textbf{Causal Inference in LLMs}: Recent work has investigated whether LLMs can perform causal inference \cite{causalInferenceLLMs, CausalInferenceLLM2, CausalInferenceLLMBench}. In this line of research, the goal is to evaluate causal-effect judgments that depend on an underlying causal structure, for example, assessing whether a model can reason about causal links under different structural assumptions \cite{cladder}. A common formulation is to pose causal questions in natural language while assuming a causal model behind the question \cite{cladder, causalInferenceLLMs}. Prior work also explores how to elicit causal inference behavior from LLMs, including causal graphs and procedural guidance that implements parts of a causal inference process \cite{cladder,causalInferenceLLMs}. In this research context, our work complements prior studies on causal inference with LLMs in two ways. First, we introduce a training strategy that aims to reduce LLMs' reliance on semantics in causality-related tasks. Second, we propose a benchmark for evaluating models’ causal reasoning ability: It formulates causal judgment questions over event triples, constructs semantically similar question pairs with opposite labels, and uses a pairwise train-test split so that semantic matching is systematically penalized.

\noindent\textbf{Causal Benchmarks for LLMs}: Prior work has proposed a range of benchmarks to evaluate causal understanding in LLMs: \textit{(i) Commonsense cause-effect} benchmarks test whether models can select plausible cause or effect under everyday knowledge \cite{COPA}. \textit{(ii) Causality-text comprehension} benchmark tests whether models can apply causal knowledge from a passage to a new situation \cite{ropesCausal}. \textit{(iii) Graph-grounded causal inference} benchmark evaluate model's causal judgment using causal graphs and formal causal questions \cite{cladder}. \textit{(iv) Comprehensive causal reasoning} benchmark provides broader causal reasoning coverage across domains such as texts, math, etc., \cite{CausalInferenceLLMBench}. Finally, \textit{(v) general LLM evaluation} benchmarks, such as BIG-bench, include causality-adjacent reasoning tasks and are often used to report general reasoning capabilities \cite{BigBench}.
However, such benchmark accuracy can be an unreliable metric for true causal reasoning ability, due to the consistent finding that models can achieve high accuracy by exploiting spurious correlations with labels rather than reasoning from the underlying causal structure \cite{Shortcuts1, Shortcuts2, SpuriousBERT}. A related concern, the reliance on spurious correlations under distribution shifts, has also been studied in other LLM applications, such as a state-of-the-art graph large language model for out-of-distribution drug synergy prediction~\cite{wang2026oodgraphllm}. CausalFlip is designed to address this issue by using semantically similar, label-flipped pairs under a pairwise train-test split. Models that rely on spurious semantic correlations could only achieve limited performance; therefore, correct predictions need to be grounded in causal structure.

\vspace{-2mm}
\section{Conclusion}
\label{sec:7}
In this work, we address a key gap in evaluating LLM causal judgment, where standard autoregressive trained models can lean on label-correlated semantic patterns to answer causal questions, instead of grounding decisions in an underlying causal structure. To bridge this gap, we introduce \textbf{CausalFlip}, a causality-based benchmark covering confounder, chain, and collider causal structures. Empirically, we find that strategies that heavily depend on semantic patterns fail on our benchmark, whereas supervision on intermediate causal reasoning steps improves accuracy. More importantly, the proposed implicit causal reasoning strategy shows clear strengths by progressively masking intermediate causal reasoning step tokens from supervision to reduce the model's reliance on possible spurious semantic correlation in explicitly generated intermediate reasoning steps.

\vspace{-2mm}
\section{Ethics and Fairness}
CausalFlip does not collect user data or involve human subjects, and it contains no private records or consent-required sources, such as identifiable information, medical or financial records, etc. During dataset construction, we manually inspected the data to mitigate risks of including sensitive information, offensive language, or potentially harmful stereotypes and biases in the text. We apply this review consistently throughout the full dataset to ensure that the released benchmark avoids such content to some extent. Finally, this benchmark is intended for research evaluation purposes, and we encourage responsible use, including clear communication of limitations, and adherence to established ethical guidelines when developing downstream applications.

\vspace{-1mm}
\section*{Acknowledgements}
This work was supported in part by the National Science Foundation (NSF) under Grants IIS-2144209, IIS-2223769, BCS-2228534, and CMMI-2411248, and by the Office of Naval Research (ONR) under Grant N000142412636.


\bibliographystyle{ACM-Reference-Format}
\bibliography{sample-base}

@misc{deng2024implicitcot,
      title={From Explicit CoT to Implicit CoT: Learning to Internalize CoT Step by Step}, 
      author={Yuntian Deng and Yejin Choi and Stuart Shieber},
      year={2024},
      eprint={2405.14838},
      archivePrefix={arXiv},
      primaryClass={cs.CL},
      url={https://arxiv.org/abs/2405.14838}, 
}

@misc{ApplyInMedicalUse,
      title={Large Language Models Encode Clinical Knowledge}, 
      author={Karan Singhal and Shekoofeh Azizi and Tao Tu and S. Sara Mahdavi and Jason Wei and Hyung Won Chung and Nathan Scales and Ajay Tanwani and Heather Cole-Lewis and Stephen Pfohl and Perry Payne and Martin Seneviratne and Paul Gamble and Chris Kelly and Nathaneal Scharli and Aakanksha Chowdhery and Philip Mansfield and Blaise Aguera y Arcas and Dale Webster and Greg S. Corrado and Yossi Matias and Katherine Chou and Juraj Gottweis and Nenad Tomasev and Yun Liu and Alvin Rajkomar and Joelle Barral and Christopher Semturs and Alan Karthikesalingam and Vivek Natarajan},
      year={2022},
      eprint={2212.13138},
      archivePrefix={arXiv},
      primaryClass={cs.CL},
      url={https://arxiv.org/abs/2212.13138}, 
}

@misc{ApplyInMedicalUse2,
      title={Towards Accurate Differential Diagnosis with Large Language Models}, 
      author={Daniel McDuff and Mike Schaekermann and Tao Tu and Anil Palepu and Amy Wang and Jake Garrison and Karan Singhal and Yash Sharma and Shekoofeh Azizi and Kavita Kulkarni and Le Hou and Yong Cheng and Yun Liu and S Sara Mahdavi and Sushant Prakash and Anupam Pathak and Christopher Semturs and Shwetak Patel and Dale R Webster and Ewa Dominowska and Juraj Gottweis and Joelle Barral and Katherine Chou and Greg S Corrado and Yossi Matias and Jake Sunshine and Alan Karthikesalingam and Vivek Natarajan},
      year={2023},
      eprint={2312.00164},
      archivePrefix={arXiv},
      primaryClass={cs.CY},
      url={https://arxiv.org/abs/2312.00164}, 
}

@misc{ApplyInFinance1,
      title={BloombergGPT: A Large Language Model for Finance}, 
      author={Shijie Wu and Ozan Irsoy and Steven Lu and Vadim Dabravolski and Mark Dredze and Sebastian Gehrmann and Prabhanjan Kambadur and David Rosenberg and Gideon Mann},
      year={2023},
      eprint={2303.17564},
      archivePrefix={arXiv},
      primaryClass={cs.LG},
      url={https://arxiv.org/abs/2303.17564}, 
}

@misc{ApplyInFinance2,
      title={FinGPT: Open-Source Financial Large Language Models}, 
      author={Hongyang Yang and Xiao-Yang Liu and Christina Dan Wang},
      year={2025},
      eprint={2306.06031},
      archivePrefix={arXiv},
      primaryClass={q-fin.ST},
      url={https://arxiv.org/abs/2306.06031}, 
}

@article{ApplyInLegal,
title = {Large language models in law: A survey},
journal = {AI Open},
volume = {5},
pages = {181-196},
year = {2024},
issn = {2666-6510},
doi = {https://doi.org/10.1016/j.aiopen.2024.09.002},
url = {https://www.sciencedirect.com/science/article/pii/S2666651024000172},
author = {Jinqi Lai and Wensheng Gan and Jiayang Wu and Zhenlian Qi and Philip S. Yu}
}

@inproceedings{COPA,
	address = {Stanford University},
	title = {Choice of {Plausible} {Alternatives}: {An} {Evaluation} of {Commonsense} {Causal} {Reasoning}},
	url = {http://ict.usc.edu/pubs/Choice%20of%20Plausible%20Alternatives-%20An%20Evaluation%20of%20Commonsense%20Causal%20Reasoning.pdf},
	booktitle = {{AAAI} {Spring} {Symposium} on {Logical} {Formalizations} of {Commonsense} {Reasoning}},
	author = {Roemmele, Melissa and Bejan, Cosmin Adrian and Gordon, Andrew S.},
	month = mar,
	year = {2011}
}

@misc{BigBench,
      title={Beyond the Imitation Game: Quantifying and extrapolating the capabilities of language models}, 
      author={Aarohi Srivastava and Abhinav Rastogi and Abhishek Rao and others},
      year={2023},
      eprint={2206.04615},
      archivePrefix={arXiv},
      primaryClass={cs.CL},
      url={https://arxiv.org/abs/2206.04615}, 
}

@misc{BigBenchHard,
      title={Challenging BIG-Bench Tasks and Whether Chain-of-Thought Can Solve Them}, 
      author={Mirac Suzgun and Nathan Scales and Nathanael Schärli and Sebastian Gehrmann and Yi Tay and Hyung Won Chung and Aakanksha Chowdhery and Quoc V. Le and Ed H. Chi and Denny Zhou and Jason Wei},
      year={2022},
      eprint={2210.09261},
      archivePrefix={arXiv},
      primaryClass={cs.CL},
      url={https://arxiv.org/abs/2210.09261}, 
}

@misc{ropesCausal,
      title={Reasoning Over Paragraph Effects in Situations}, 
      author={Kevin Lin and Oyvind Tafjord and Peter Clark and Matt Gardner},
      year={2019},
      eprint={1908.05852},
      archivePrefix={arXiv},
      primaryClass={cs.CL},
      url={https://arxiv.org/abs/1908.05852}, 
}

@inproceedings{AutoRGPT1,
  title={Improving Language Understanding by Generative Pre-Training},
  author={Alec Radford and Karthik Narasimhan},
  year={2018},
  url={https://api.semanticscholar.org/CorpusID:49313245}
}

@misc{AutoRGPT3,
      title={Language Models are Few-Shot Learners}, 
      author={Tom B. Brown and Benjamin Mann and Nick Ryder and Melanie Subbiah and Jared Kaplan and Prafulla Dhariwal and Arvind Neelakantan and Pranav Shyam and Girish Sastry and Amanda Askell and Sandhini Agarwal and Ariel Herbert-Voss and Gretchen Krueger and Tom Henighan and Rewon Child and Aditya Ramesh and Daniel M. Ziegler and Jeffrey Wu and Clemens Winter and Christopher Hesse and Mark Chen and Eric Sigler and Mateusz Litwin and Scott Gray and Benjamin Chess and Jack Clark and Christopher Berner and Sam McCandlish and Alec Radford and Ilya Sutskever and Dario Amodei},
      year={2020},
      eprint={2005.14165},
      archivePrefix={arXiv},
      primaryClass={cs.CL},
      url={https://arxiv.org/abs/2005.14165}, 
}

@inproceedings{Shortcuts1,
   title={Large Language Models Can be Lazy Learners: Analyze Shortcuts in In-Context Learning},
   url={http://dx.doi.org/10.18653/v1/2023.findings-acl.284},
   DOI={10.18653/v1/2023.findings-acl.284},
   booktitle={Findings of the Association for Computational Linguistics: ACL 2023},
   publisher={Association for Computational Linguistics},
   author={Tang, Ruixiang and Kong, Dehan and Huang, Longtao and Xue, Hui},
   year={2023},
   pages={4645–4657} }

@misc{Shortcuts2,
      title={Why Machine Reading Comprehension Models Learn Shortcuts?}, 
      author={Yuxuan Lai and Chen Zhang and Yansong Feng and Quzhe Huang and Dongyan Zhao},
      year={2021},
      eprint={2106.01024},
      archivePrefix={arXiv},
      primaryClass={cs.CL},
      url={https://arxiv.org/abs/2106.01024}, 
}

@misc{SpuriousBERT,
      title={Right for the Wrong Reasons: Diagnosing Syntactic Heuristics in Natural Language Inference}, 
      author={R. Thomas McCoy and Ellie Pavlick and Tal Linzen},
      year={2019},
      eprint={1902.01007},
      archivePrefix={arXiv},
      primaryClass={cs.CL},
      url={https://arxiv.org/abs/1902.01007}, 
}

@misc{ChainOfThoughtWei,
      title={Chain-of-Thought Prompting Elicits Reasoning in Large Language Models}, 
      author={Jason Wei and Xuezhi Wang and Dale Schuurmans and Maarten Bosma and Brian Ichter and Fei Xia and Ed Chi and Quoc Le and Denny Zhou},
      year={2023},
      eprint={2201.11903},
      archivePrefix={arXiv},
      primaryClass={cs.CL},
      url={https://arxiv.org/abs/2201.11903}, 
}

@misc{ChainOfThoughtZeroshot,
      title={Large Language Models are Zero-Shot Reasoners}, 
      author={Takeshi Kojima and Shixiang Shane Gu and Machel Reid and Yutaka Matsuo and Yusuke Iwasawa},
      year={2023},
      eprint={2205.11916},
      archivePrefix={arXiv},
      primaryClass={cs.CL},
      url={https://arxiv.org/abs/2205.11916}, 
}

@inproceedings{CoTCost,
    title = "{S}pec{C}o{T}: Accelerating Chain-of-Thought Reasoning through Speculative Exploration",
    author = "Shi, Junhan  and
      Zhu, Yijia  and
      Shi, Zhenning  and
      Zhao, Dan  and
      Li, Qing  and
      Jiang, Yong",
    editor = "Christodoulopoulos, Christos  and
      Chakraborty, Tanmoy  and
      Rose, Carolyn  and
      Peng, Violet",
    booktitle = "Findings of the Association for Computational Linguistics: EMNLP 2025",
    month = nov,
    year = "2025",
    address = "Suzhou, China",
    publisher = "Association for Computational Linguistics",
    url = "https://aclanthology.org/2025.findings-emnlp.1326/",
    doi = "10.18653/v1/2025.findings-emnlp.1326",
    pages = "24405--24415",
    ISBN = "979-8-89176-335-7"
}

@misc{CoTCost2,
      title={Compressed Chain of Thought: Efficient Reasoning Through Dense Representations}, 
      author={Jeffrey Cheng and Benjamin Van Durme},
      year={2024},
      eprint={2412.13171},
      archivePrefix={arXiv},
      primaryClass={cs.CL},
      url={https://arxiv.org/abs/2412.13171}, 
}

@misc{Attention,
      title={Attention Is All You Need}, 
      author={Ashish Vaswani and Noam Shazeer and Niki Parmar and Jakob Uszkoreit and Llion Jones and Aidan N. Gomez and Lukasz Kaiser and Illia Polosukhin},
      year={2023},
      eprint={1706.03762},
      archivePrefix={arXiv},
      primaryClass={cs.CL},
      url={https://arxiv.org/abs/1706.03762}, 
}

@article{ProbLM,
author = {Bengio, Yoshua and Ducharme, R\'{e}jean and Vincent, Pascal and Janvin, Christian},
title = {A neural probabilistic language model},
year = {2003},
issue_date = {3/1/2003},
publisher = {JMLR.org},
volume = {3},
number = {null},
issn = {1532-4435},
journal = {J. Mach. Learn. Res.},
month = mar,
pages = {1137–1155},
numpages = {19}
}

@misc{AutoLimitaions,
      title={Faith and Fate: Limits of Transformers on Compositionality}, 
      author={Nouha Dziri and Ximing Lu and Melanie Sclar and Xiang Lorraine Li and Liwei Jiang and Bill Yuchen Lin and Peter West and Chandra Bhagavatula and Ronan Le Bras and Jena D. Hwang and Soumya Sanyal and Sean Welleck and Xiang Ren and Allyson Ettinger and Zaid Harchaoui and Yejin Choi},
      year={2023},
      eprint={2305.18654},
      archivePrefix={arXiv},
      primaryClass={cs.CL},
      url={https://arxiv.org/abs/2305.18654}, 
}

@misc{sqtosq,
      title={Sequence to Sequence Learning with Neural Networks}, 
      author={Ilya Sutskever and Oriol Vinyals and Quoc V. Le},
      year={2014},
      eprint={1409.3215},
      archivePrefix={arXiv},
      primaryClass={cs.CL},
      url={https://arxiv.org/abs/1409.3215}, 
}

@misc{topP,
      title={The Curious Case of Neural Text Degeneration}, 
      author={Ari Holtzman and Jan Buys and Li Du and Maxwell Forbes and Yejin Choi},
      year={2020},
      eprint={1904.09751},
      archivePrefix={arXiv},
      primaryClass={cs.CL},
      url={https://arxiv.org/abs/1904.09751}, 
}

@misc{topK,
      title={Hierarchical Neural Story Generation}, 
      author={Angela Fan and Mike Lewis and Yann Dauphin},
      year={2018},
      eprint={1805.04833},
      archivePrefix={arXiv},
      primaryClass={cs.CL},
      url={https://arxiv.org/abs/1805.04833}, 
}

@misc{causalInferenceLLMs,
      title={Causal Reasoning and Large Language Models: Opening a New Frontier for Causality}, 
      author={Emre Kıcıman and Robert Ness and Amit Sharma and Chenhao Tan},
      year={2024},
      eprint={2305.00050},
      archivePrefix={arXiv},
      primaryClass={cs.AI},
      url={https://arxiv.org/abs/2305.00050}, 
}

@misc{cladder,
      title={CLadder: Assessing Causal Reasoning in Language Models}, 
      author={Zhijing Jin and Yuen Chen and Felix Leeb and Luigi Gresele and Ojasv Kamal and Zhiheng Lyu and Kevin Blin and Fernando Gonzalez Adauto and Max Kleiman-Weiner and Mrinmaya Sachan and Bernhard Schölkopf},
      year={2024},
      eprint={2312.04350},
      archivePrefix={arXiv},
      primaryClass={cs.CL},
      url={https://arxiv.org/abs/2312.04350}, 
}

@inproceedings{CausalInferenceLLMBench,
    title = "{C}ausal{B}ench: A Comprehensive Benchmark for Evaluating Causal Reasoning Capabilities of Large Language Models",
    author = "Wang, Zeyu",
    editor = "Wong, Kam-Fai  and
      Zhang, Min  and
      Xu, Ruifeng  and
      Li, Jing  and
      Wei, Zhongyu  and
      Gui, Lin  and
      Liang, Bin  and
      Zhao, Runcong",
    booktitle = "Proceedings of the 10th SIGHAN Workshop on Chinese Language Processing (SIGHAN-10)",
    month = aug,
    year = "2024",
    address = "Bangkok, Thailand",
    publisher = "Association for Computational Linguistics",
    url = "https://aclanthology.org/2024.sighan-1.17/",
    pages = "143--151"
}

@misc{CausalInferenceLLM2,
      title={Can Large Language Models Infer Causation from Correlation?}, 
      author={Zhijing Jin and Jiarui Liu and Zhiheng Lyu and Spencer Poff and Mrinmaya Sachan and Rada Mihalcea and Mona Diab and Bernhard Schölkopf},
      year={2024},
      eprint={2306.05836},
      archivePrefix={arXiv},
      primaryClass={cs.CL},
      url={https://arxiv.org/abs/2306.05836}, 
}

@misc{hallucination1,
      title={On Faithfulness and Factuality in Abstractive Summarization}, 
      author={Joshua Maynez and Shashi Narayan and Bernd Bohnet and Ryan McDonald},
      year={2020},
      eprint={2005.00661},
      archivePrefix={arXiv},
      primaryClass={cs.CL},
      url={https://arxiv.org/abs/2005.00661}, 
}

@misc{hallucination2,
      title={Why Language Models Hallucinate}, 
      author={Adam Tauman Kalai and Ofir Nachum and Santosh S. Vempala and Edwin Zhang},
      year={2025},
      eprint={2509.04664},
      archivePrefix={arXiv},
      primaryClass={cs.CL},
      url={https://arxiv.org/abs/2509.04664}, 
}

@misc{hallucination3,
      title={TruthfulQA: Measuring How Models Mimic Human Falsehoods}, 
      author={Stephanie Lin and Jacob Hilton and Owain Evans},
      year={2022},
      eprint={2109.07958},
      archivePrefix={arXiv},
      primaryClass={cs.CL},
      url={https://arxiv.org/abs/2109.07958}, 
}

@misc{stella400M,
      title={Jasper and Stella: distillation of SOTA embedding models}, 
      author={Dun Zhang and Jiacheng Li and Ziyang Zeng and Fulong Wang},
      year={2025},
      eprint={2412.19048},
      archivePrefix={arXiv},
      primaryClass={cs.IR},
      url={https://arxiv.org/abs/2412.19048}, 
}

@misc{lora,
      title={LoRA: Low-Rank Adaptation of Large Language Models}, 
      author={Edward J. Hu and Yelong Shen and Phillip Wallis and Zeyuan Allen-Zhu and Yuanzhi Li and Shean Wang and Lu Wang and Weizhu Chen},
      year={2021},
      eprint={2106.09685},
      archivePrefix={arXiv},
      primaryClass={cs.CL},
      url={https://arxiv.org/abs/2106.09685}, 
}

@misc{adamW,
      title={Decoupled Weight Decay Regularization}, 
      author={Ilya Loshchilov and Frank Hutter},
      year={2019},
      eprint={1711.05101},
      archivePrefix={arXiv},
      primaryClass={cs.LG},
      url={https://arxiv.org/abs/1711.05101}, 
}

@misc{gradientcheckpointing,
      title={Training Deep Nets with Sublinear Memory Cost}, 
      author={Tianqi Chen and Bing Xu and Chiyuan Zhang and Carlos Guestrin},
      year={2016},
      eprint={1604.06174},
      archivePrefix={arXiv},
      primaryClass={cs.LG},
      url={https://arxiv.org/abs/1604.06174}, 
}

@misc{bf16,
      title={A Study of BFLOAT16 for Deep Learning Training}, 
      author={Dhiraj Kalamkar and Dheevatsa Mudigere and Naveen Mellempudi and others},
      year={2019},
      eprint={1905.12322},
      archivePrefix={arXiv},
      primaryClass={cs.LG},
      url={https://arxiv.org/abs/1905.12322}, 
}

@article{wang2026oodgraphllm,
  title={OOD-GraphLLM: Graph Large Language Model for Out-of-Distribution Generalized Drug Synergy Prediction},
  author={Wang, Xin and Xiao, Linxin and Yao, Yang and Zhu, Wenwu},
  journal={arXiv preprint arXiv:2605.30247},
  year={2026}
}

\appendix
\section{Appendix}
\subsection{Reducing Data Skewness}
Although the pairwise causal questions with opposite causal answers are specifically designed to be semantically close to each other, the dataset as a whole can still contain skewed correlations between semantic features and labels that LLMs could learn as spurious shortcuts \cite{Shortcuts1, Shortcuts2}. Here, ``skewness'' refers to distributional imbalances where particular events appear disproportionately in certain categories. For example, certain events can still show up more in one category than the others, such as sales of certain products or certain weather conditions appearing more as events in Base pairs than Opposite pairs. Such imbalances can create shortcut signals that allow models to predict labels without grounding the reasoning in causal structure.

We detect such skewness from two perspectives. First, we analyze the \emph{count} of events in each dataset, computing their frequencies across labels and categories and flagging those with highly imbalanced distributions under a threshold. In addition, we perform \emph{similarity-based} analysis to identify examples that dominate the dataset's semantic space. Specifically, we embed each question using the \texttt{stella-400M} model \cite{stella400M} and compute cosine similarity; for each example, we retrieve its top-5 nearest neighbors in the dataset. We then count how often an example appears in the top-5 neighbor lists of other examples. Examples that appear unusually frequently in these top-5 lists indicate near-duplicate clusters or highly generic patterns that may amplify label-correlated shortcuts.

To mitigate this skewness, we filter and replace the top 5 cases identified by the two analyses iteratively. For count-based skew, we replace overrepresented events (in $X$, $Y$, or $Z$) with substitute events so that event frequencies are more evenly distributed across categories. For similarity-based analysis, we replace examples that appear unusually often as a top-5 neighbor of many others, reducing skewness induced by similarity and preventing highly repetitive patterns from dominating the training shortcuts.

\vspace{-1mm}
\subsection{Hyperparameters}
For all fine-tuned strategies (no-CoT, explicit-CoT and implicit causal reasoning), we apply Low-Rank Adaptation (LoRA) \cite{lora} with $r=4$, $\alpha=8$, dropout $=0.05$, targeting \textit{q\_proj}, \textit{v\_proj}. We fine-tune for 3 epochs with learning rate $1\times 10^{-4}$, batch size 4, maximum sequence length 256, using the paged AdamW \cite{adamW} 8-bit optimizer, bf16 training \cite{bf16}, and gradient checkpointing \cite{gradientcheckpointing}, with fixed random seed for fair comparisons.

\vspace{-1mm}
\subsection{Benchmark Generalizability}
\textbf{CausalFlip} exposes vulnerabilities from models' reliance on spurious semantic correlations. To further examine whether the behavior observed on CausalFlip is specific to the Llama-3.2-3B-Instruct used in the main experiments, we additionally evaluate IBM-PowerMoE-3B, a Mixture-of-Experts model with a different architecture from the dense transformer backbone used in the primary experiments. This experiment served as a \textbf{controlled cross-architecture check}.
\begin{table}[t]
\centering
\caption{IBM-PowerMoE-3B results on CausalFlip. Standard fine-tuning has lower accuracy than the pretrained baseline across all three sub-datasets.}
\vspace{-2mm}
\label{tab:ibm_result}
\resizebox{\columnwidth}{!}{%
\begin{tabular}{lcccc}
\toprule
\textbf{Dataset} & \textbf{Pretrained} & \textbf{Valid Yes/No} & \textbf{No CoT} & \textbf{Valid Yes/No} \\
\midrule
Confounder & 0.519 & 1000 / 1000 & 0.508 & 999 / 1000 \\
Chain      & 0.508 & 1000 / 1000 & 0.500 & 993 / 1000 \\
Collider   & 0.532 & 1000 / 1000 & 0.528 & 995 / 1000 \\
\bottomrule
\end{tabular}
}
\vspace{-1mm}
\end{table}
Table~\ref{tab:ibm_result} reports the results of IBM-PowerMoE-3B under the naive pretraining strategy and standard no-CoT fine-tuning. Across all three sub-datasets, it shows \textbf{similar vulnerability}. Standard fine-tuning decreases performance compared to the pretrained baseline. Accuracy drops from 51.9\% to 50.8\% on Confounder, from 50.8\% to 50.0\% on Chain, and from 53.2\% to 52.8\% on Collider. Valid-output rates are high in both settings: 1000/1000 for the pretrained model on all three datasets, and 999/1000, 993/1000, and 995/1000 for the fine-tuned model on Confounder, Chain, and Collider. Thus, the observed result is not driven by invalid output.

These results suggest that the vulnerabilities identified by our benchmark are not restricted to the Llama-based model. Even with a different Mixture-of-Experts architecture, the model could still rely on spurious semantic correlations, causing standard fine-tuning to fail to improve causal judgment under our label-flipped, semantically similar question-pair benchmark setting. This supports the role of CausalFlip as a benchmark for detecting reliance on spurious semantic correlations across model architectures.
\begin{table}[t]
\centering
\caption{Representation layer comparisons. We compare the last-layer hidden representation at the final prompt token across no-CoT, explicit-CoT, and implicit causal reasoning.}
\vspace{-2mm}
\label{tab:representation}
\resizebox{\columnwidth}{!}{%
\begin{tabular}{llcc}
\toprule
\textbf{Dataset} & \textbf{Comparison} & \textbf{Cosine Similarity} & \textbf{Linear CKA} \\
\midrule
Confounder & Implicit vs. Explicit CoT & 0.8928 & 0.9890 \\
Confounder & Implicit vs. No CoT       & 0.1174 & 0.4757 \\
Confounder & Explicit CoT vs. No CoT   & 0.1198 & 0.4933 \\
\midrule
Chain & Implicit vs. Explicit CoT & 0.7263 & 0.9038 \\
Chain & Implicit vs. No CoT       & 0.1364 & 0.5376 \\
Chain & Explicit CoT vs. No CoT   & 0.1203 & 0.3802 \\
\midrule
Collider & Implicit vs. Explicit CoT & 0.8146 & 0.9431 \\
Collider & Implicit vs. No CoT       & 0.0905 & 0.4429 \\
Collider & Explicit CoT vs. No CoT   & 0.0795 & 0.3858 \\
\bottomrule
\end{tabular}
}
\vspace{-10pt}
\end{table}

\subsection{Interpretability in Internalization}

To provide additional evidence for the internalization of implicit causal reasoning, we conduct a representation-level similarity comparison among no-CoT, explicit-CoT, and implicit causal reasoning. For each trained model, we feed the same question prompt and extract the final-layer hidden representation at the last prompt token, before the model begins generating the answer. We then compare the representations across methods using cosine similarity and linear centered kernel alignment (CKA).

Table~\ref{tab:representation} reports the representation similarity on the test sets of Confounder, Chain, and Collider. Across all sub-datasets, implicit causal reasoning consistently achieves high similarity to explicit-CoT, and is much more similar to explicit-CoT than to no-CoT. Specifically, for cosine similarity, implicit causal reasoning is more similar to explicit-CoT than to no-CoT on Confounder (0.8928 vs. 0.1174), Chain (0.7263 vs. 0.1364), and Collider (0.8146 vs. 0.0905). Similarly, for linear CKA, implicit causal reasoning has higher similarity to explicit-CoT than to no-CoT on Confounder (0.9890 vs. 0.4757), Chain (0.9038 vs. 0.5376), and Collider (0.9431 vs. 0.4429).

Therefore, these results provide complementary representation-level evidence for the internalization from progressive masking. The representations under implicit causal reasoning are highly similar to those under explicit-CoT, and are much more similar to explicit-CoT than to those under no-CoT. This supports the interpretation that implicit causal reasoning indeed internalizes CoT-like reasoning.
\end{document}